\documentclass[11pt]{article}

\usepackage{microtype} 
\usepackage{booktabs}  
\usepackage{url}  
\usepackage{csquotes}

\usepackage{xcolor}

\colorlet{azcolor}{pink}
\colorlet{dpcolor}{green!80}
\colorlet{kwcolor}{violet!80}
\colorlet{akcolor}{blue!80}

\usepackage{paralist}

\usepackage{amsmath}
\usepackage{amsthm}
\usepackage{bbm}

\usepackage{caption}
\usepackage{subcaption}
\usepackage{graphicx}

\usepackage{hyperref}
\usepackage{booktabs}
\usepackage{longtable}
\usepackage{multirow}

\usepackage[workshop, final]{automl}
\usepackage{xcolor}
\usepackage{natbib}
\usepackage{filecontents}

\newtheorem{assumptionex}{Requirement}
\newenvironment{assumption}
  {\pushQED\renewcommand\assumptionex}
  {\popQED\endassumptionex}

\title{Rethinking of Encoder-based Warm-start Methods in Hyperparameter Optimization}

\author[1,$\ast$]{\nameemail{Dawid Płudowski}{dawid.pludowski@gmail.com}}
\author[1,$\ast$]{\nameemail{Antoni Zajko}{antoni.zajko.1@gmail.com}}
\author[1]{\nameemail{Anna Kozak}{anna.kozak@pw.edu.pl}}
\author[1]{\nameemail{Katarzyna Woźnica}{katarzyna.woznica@pw.edu.pl}}

\affil[$\ast$]{Equal contribution.}
\affil[1]{Warsaw University of Technology}

\hypersetup{%
  pdfauthor={AutoML}, 
  pdftitle={Rethinking of Encoder-Based Warmstart Methods in Hyperparameter Optimization},
  pdfsubject={Rethinking of Encoder-Based Warmstart Methods in Hyperparameter Optimization},
  pdfkeywords={AutoML, LaTeX, style}
}

\begin{document}

\maketitle

\begin{abstract}

Effectively representing heterogeneous tabular datasets for meta-learning purposes remains an open problem. Previous approaches rely on predefined meta-features, for example, statistical measures or landmarkers. The emergence of dataset encoders opens new possibilities for the extraction of meta-features because they do not involve any handmade design. Moreover, they are proven to generate dataset representations with desired spatial properties. In this research, we evaluate an encoder-based approach to one of the most established meta-tasks -- warm-starting of the Bayesian Hyperparameter Optimization. To broaden our analysis we introduce a new approach for representation learning on tabular data based on~\citep{iwata_meta-learning_2020}. The validation on over 100 datasets from UCI and an independent metaMIMIC set of datasets highlights the nuanced challenges in representation learning. We show that general representations may not suffice for some meta-tasks where requirements are not explicitly considered during extraction.
 
\end{abstract}

\section{Introduction}

The meta-learning for tabular data poses a unique challenge for the machine learning community as there is no universal way to compare two datasets. Yet, such data remains ubiquitous across various domains~\citep{davenport_potential_2019,alanazi_using_2022}. To resolve this issue, new methods for data representation were developed~\citep {vanschoren_meta-learning_2018,jomaa_dataset2vec_2021}.

\vspace{0.2cm}

\noindent \textbf{Problem motivation.} So far existing approaches to datasets' representation rely primarily on handcrafted meta-features which are often based on statistical measures, information theory, or landmarkers~\citep{rivolli_meta-features_2022}. The emergence of encoders tailored to heterogeneous datasets solved this problem by limiting the need for manually chosen meta-features since they only require a loss function that formalizes the desired properties of the representations. One of the first encoder-based approaches was Dataset2Vec~\citep{jomaa_dataset2vec_2021} which introduced:

\begin{assumption}
\label{ass:similarity}
Ensuring close representations for batches of observations within the same dataset while maintaining distinct representations for observations across different datasets. 
\end{assumption}

\noindent By imposing this condition, the authors assume that the resulting representations will reflect the internal complex structure of the data. 
This is, however, not explicit whether such representations are useful in meta-learning tasks, like Bayesian Optimization (BO).

\vspace{0.2cm}
\noindent
\textbf{Contributions.} In this work, we:
\begin{inparaenum}[(1)]
    \item conducted a comprehensive analysis of encoder-based representations using two distinct sets of datasets on the BO warm-start problem,
    \item proposed new dataset encoder which is rooted in Requirement \ref{ass:similarity} and inspired by the few-shot architecture introduced in~\citep{iwata_meta-learning_2020}.  
\end{inparaenum}

The validation of encoder-based representations' applicability in meta-tasks highlights the nuanced challenges in representation learning. \textbf{Merely adhering to Requirement~\ref{ass:similarity} through encoder-based representations proves to be inadequate for broader meta-learning.}

\section{Problem setting}
\label{sec:def}

\begin{figure}[!h]
     \centering
     \includegraphics[width=\textwidth]{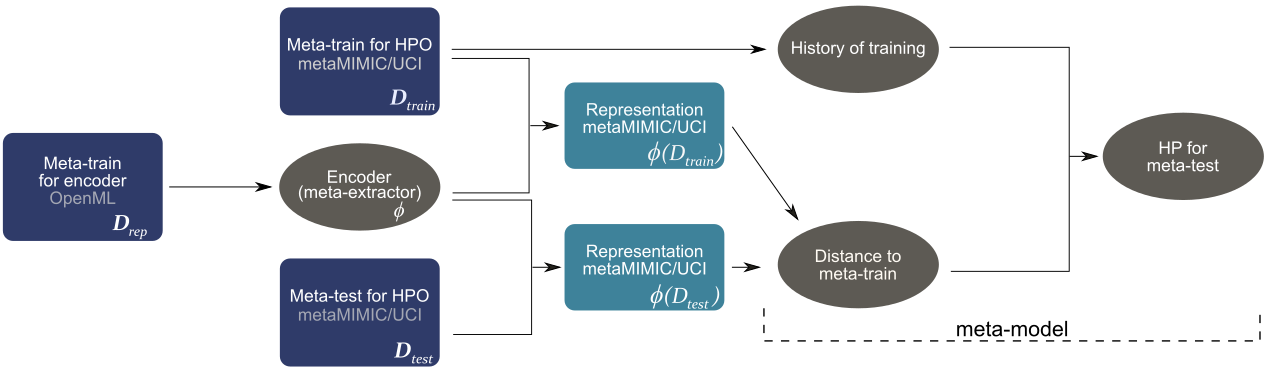}
     \caption{The workflow of the performed experiment. We denote datasets using dark blue color, their representations using light blue, and all other components using grey.}
     \label{img:methodology-graph}

\end{figure}

In this section, we provide the formulation of the problem of employing dataset representation in meta-learning tasks. In Figure~\ref{img:methodology-graph}, we show the defined components and their workflow. First, we train encoder $\phi$ on OpenML's subset $\mathbfcal{D}_{rep}$. The encoder is either \textit{liltab} or D2V. Next, we create representations of datasets from the train and test set $\mathbfcal{D}_{train}$, $\mathbfcal{D}_{test}$. The origin of these datasets is either the UCI repository or metaMIMIC, depending on the experiment setting. In the next step, we create a mapping between the datasets from $\mathbfcal{D}_{train}$ and the corresponding best configurations of hyperparameters (HP) for the predictive algorithm, elastic net, and XGBoost. Having this mapping and the distances from a single dataset from test set $\mathcal{D}_{i}$ to datasets from $\mathbfcal{D}_{train}$, we run BO with warm-start chosen among best configurations for 10 closest datasets from $\mathbfcal{D}_{train}$.

As shown in the Figure \ref{img:methodology-graph}, our experiment setting follows a two-step process of solving meta-tasks provided in~\citep{jomaa_dataset2vec_2021}:  
\begin{inparaenum}[(1)]
    \item finding the representation in the common vector space  of heterogeneous tabular datasets,
    \item using representation as an input to the meta-model which solves the meta-task.
\end{inparaenum}

We use the BO warm-start in our experiments due to its popularity in research \citep{james_bergstra_algorithms_2011, nicoletta_del_buono_methods_2020, xiyuan_liu_efficient_2020, a_helen_victoria_automatic_2021} and its significance in the speeding up the BO process. The obtained results are compared with baselines that do not use a data structure to propose a candidate for warm-start.

\section{Considered approaches}

Next to the Dataset2Vec, we use \textit{liltab}-based encoder which meets Requirement \ref{ass:similarity}. Both work on heterogeneous data, which is crucial when working with tabular data. Their performance is contrasted with two baseline methods independent of dataset representations.

\subsection{Encoders}
Dataset2Vec (D2V) is an encoder of the whole dataset proposed in~\citep{jomaa_dataset2vec_2021}. It is one of the approaches to encoding heterogeneous tabular data that uses neural networks instead of statistical measures. 
It is a DeepSet-based  method~\citep{zaheer_deep_2017} composed of feed-forward neural networks. It extracts interdependencies between features and targets and joint distributions inside the data in a three-step process which results in fixed-size encoding.
In addition to D2V, we propose the encoder based on~\citep{iwata_meta-learning_2020}.  
The crucial feature of Iwata's architecture is the ability to perform predictions on heterogeneous tabular data in the few-shot setting. 
It is obtained by using an inference network as an encoder for the support set. In our package \textit{liltab}, we implemented both the original architecture from~\citep{iwata_meta-learning_2020} and the modified inference network as an encoder. For training purposes, we used a contrastive learning approach. For a more detailed description, see Appendix~\ref{app:liltab}.

\subsection{Baselines}

As baselines for the usability of encoder-based representations, we use two methods: BO without warm-start and warm-start based on the ranks of HP configurations. We define ranks by the number of tasks in which the HP configuration was the best. When calculating ranks we only treat the given configuration as the best one for the specific task when it is strictly the best among all considered settings i.e. there is only a single best task.

\section{Experiments}
\label{sec:experiments}

We select three sets of datasets for the experiment purposes. All of them contain only binary classification tasks: 
\begin{inparaenum}[(1)]
    \item OpenML's subset~\citep{vanschoren_openml_2013} used in~\citep{iwata_meta-learning_2020},
    \item UCI's subset~\citep{uci} used in~\citep{jomaa_dataset2vec_2021},
    \item Over 5000 few-shot tasks generated from metaMIMIC \citep{woznica_consolidated_2023} collected from~\href{https://physionet.org/}{physionet.org}.
\end{inparaenum}
Their usage is presented in Figure~\ref{img:methodology-graph}. The detailed description can be found in the Appendices~\ref{app:data} and~\ref{app:uci_data}.

We train both encoders on OpenML datasets. The details of the training process are listed in Appendix~\ref{app:encoders_hp}. Full reproduction of the training encoders, as well as the following experiments, can be done with code on GitHub\footnote{\url{https://github.com/azoz01/rethinking_encoder_warmstart}}.

We evaluate the quality of the representations in two ways: visually using T-SNE~\citep{van2008visualizing} and quantitatively using two metrics. One of them is the accuracy in the meta-task of classification whether two samples (both in terms of rows and columns) originate from the same dataset. The other one is Caliński-Harabasz (CH) index \citep{Caliński_1974} where in the place of the cluster labels we put the label of the origin of the dataset's batch. This measures how far from each other are representations depending on the origin dataset of the batch. Having these metrics, we measure the ability of encoders to distinguish between datasets and their representations' spatial properties. Therefore, they reflect meeting the Requirement~\ref{ass:similarity}. 

Next, we assess the applicability of the meta-extractors in the HPO warm-start meta-task. To achieve that we performed optimizations with encoder-based warm-start, heuristic-based warmstart, and random initialization. The gain of specific approaches was computed using Average Distance To the Minimum (ADTM) for the scaled ROC-AUC score which resulted in separate metrics for each iteration of the HPO. The experiment was performed on data from the UCI repository and metaMIMIC.  We use cross-validation concerning the selection of validation datasets. For metaMIMIC and UCI data, we use 4 folds and 5 folds respectively. Each optimization consists of 30 iterations with 10 warm-start iterations.

\subsection{Experimental setup}
\label{sec:setup}

During our research on the warm-start, we focus on searching for the best HP for the elastic net~\citep{pedregosa_scikit-learn_2018} and XGBoost~\citep{chen_xgboost_2016} classifiers. For the experiment's purpose, we generate a random set of the hyperparameters of the elastic net and XGBoost, each consisting of 100 configurations. We use XGBoost only on UCI datasets, as metaMIMIC tasks would be too small for this model. The definition of the ranges and distributions of the HP are shown in Appendix~\ref{app:ho}. Then, we train and evaluate the model with every configuration of HP on each task. For encoder-based approaches as warm-start points, we select the best configuration on the 10 datasets closest in terms of distance function $\|\cdot\|_2$. For the selection of the rest of the points, we use the Tree-structured Parzen Estimator~\citep{tpe}.

\section{Results}
This section presents two different assessment methods of tabular data representation. The first of them is the similarity of latent vectors corresponding to similar datasets, i.e., meeting Requirement~\ref{ass:similarity}. The second one is the metrics obtained from warm-start in HP optimization.

\subsection{Encodings similarity of datasets}

\begin{figure}[h]
     \centering
     \begin{subfigure}[b]{0.45\textwidth}
         \centering
         \includegraphics[width=\textwidth]{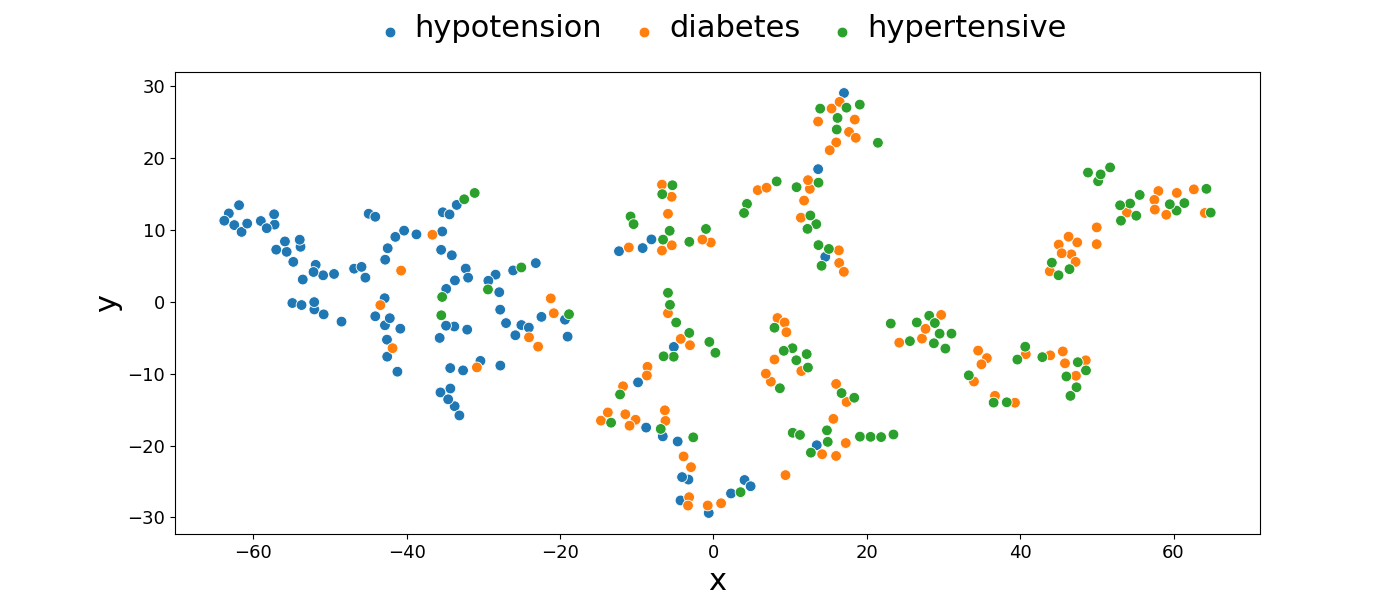}
         \caption{The representations obtained with \textit{liltab}.}
         \label{img:mimic-liltab}
     \end{subfigure}
     \hfill
     \begin{subfigure}[b]{0.45\textwidth}
         \centering
         \includegraphics[width=\textwidth]{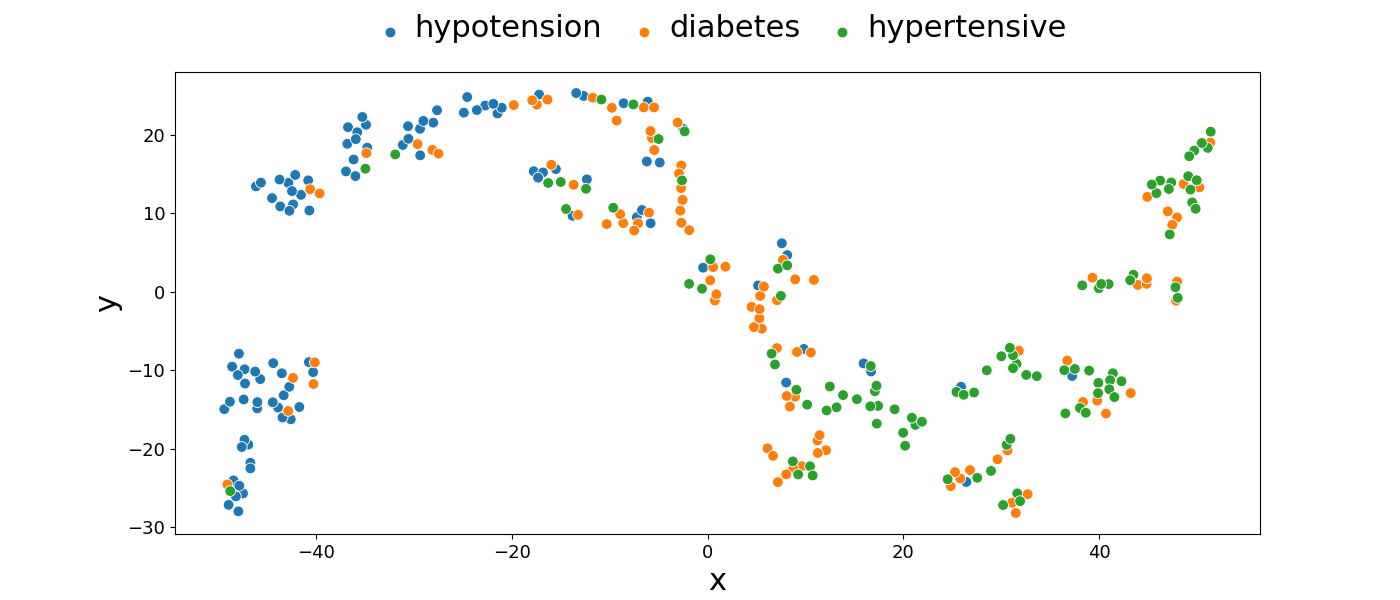}
         \caption{The representations obtained with D2V.}
         \label{img:mimic-d2v}
     \end{subfigure}
     \caption{T-SNE visualization of three encoded datasets' representations. 
     }
     \label{img:mimic}
\end{figure}

In Figure \ref{img:mimic}, we show the T-SNE representation of vectors obtained by \textit{liltab} and D2V. For this purpose, we choose three tasks from the metaMIMIC dataset where two of them - 
\texttt{hypertensive} and \texttt{diabetes} share 7 out of 10 features and third - \texttt{hypotension} which has only 3 features in common with the rest. Each point on this figure corresponds to the validation set of the single few-shot task generated from metaMIMIC. The analogous results are present for UCI datasets in Appendix \ref{app:uci}. In Table \ref{tab:encoders_metrics} we present corresponding metrics. Both encoders perform similarly, clustering the majority of datasets very well which indicates that they fulfill Requirement~\ref{ass:similarity}.

\begin{table}[h]
    \centering
    \caption{Metrics that assess the quality of representations generated by encoders. Each evaluation was repeated 15 times. We present the averaged values. Accuracy above $0.5$ indicates the ability to distinguish between datasets to some extent.}
    \label{tab:encoders_metrics}
    \begin{tabular}{llrr}
    \hline
    Encoder     & Set of datasets & Accuracy & CH index\\ \hline
    D2V & metaMIMIC           & $0.59~\pm~0.006$     &  $64.85~\pm~5.03$ \\ \hline
    \emph{liltab}      & metaMIMIC           & $0.58~\pm~0.007$     & $70.77~\pm~2.90$\\ \hline
    D2V & UCI             & $0.72~\pm~0.006$     & $241.27~\pm~5.10$ \\ \hline
    \emph{liltab}      & UCI             & $0.65~\pm~0.008$   & $136.37~\pm~8.55$   \\ \hline
    \end{tabular}
\end{table}

\subsection{Transferability of HP}

\begin{figure}
\begin{subfigure}[t]{1\textwidth}
 \centering
     \includegraphics[width=\textwidth]{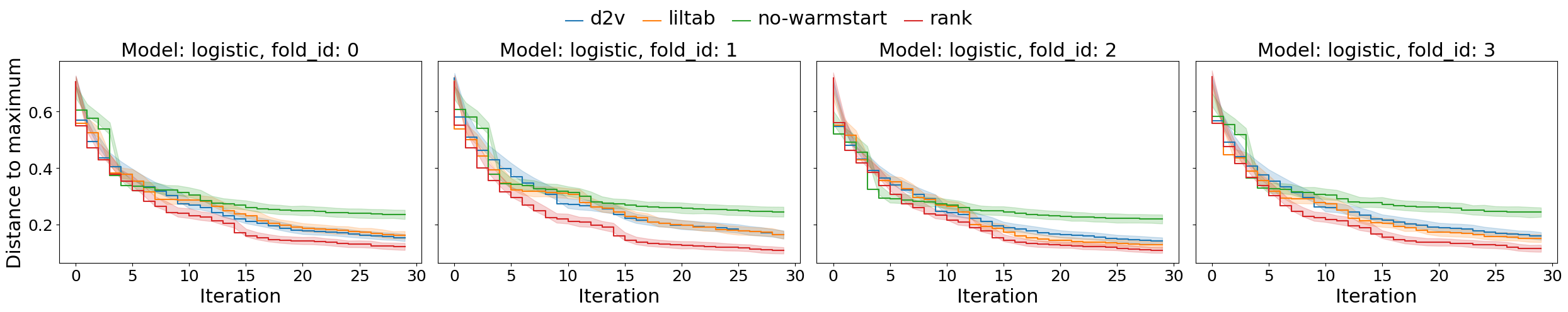}
     \caption{Plots of ADTM of metaMIMIC data HPO with warm-start comparison. A lower score means the performance of a method is closer to the best score obtained on a specific dataset on average.}
     \label{img:mimic-adtm}
\end{subfigure}
\hfill
\centering
     \begin{subfigure}[b]{0.475\textwidth}
         \centering
         \includegraphics[width=\textwidth]{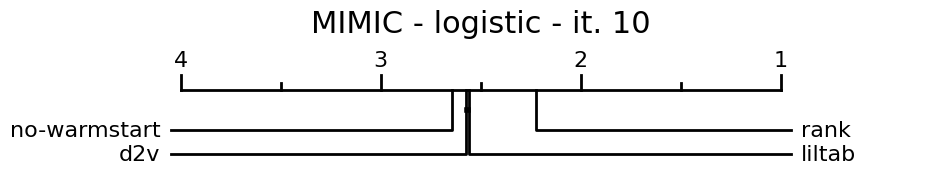}
         \caption{Critical distance plot for logistic regression in 10th iteration.}
         \label{img:mimic-cd-10}
     \end{subfigure}
     \hfill
     \begin{subfigure}[b]{0.475\textwidth}
         \centering
         \includegraphics[width=\textwidth]{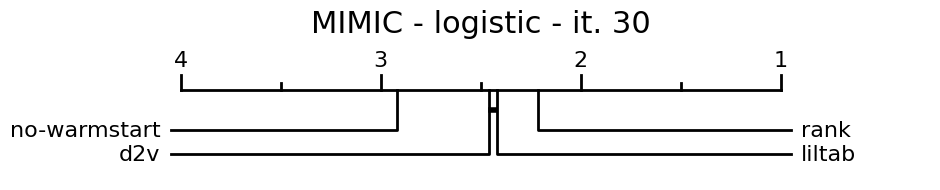}
         \caption{Critical distance plot for logistic regression in 30th iteration.}
        \label{img:mimic-cd-30}
     \end{subfigure}
     \caption{Results for the elastic net on metaMIMIC. Each fold in ADTM plots on Figure~\ref{img:mimic-adtm} presents results for a specific split of datasets to train and validate the subset. Here, the rank method results in the lowest average distance to the maximum ROC-AUC score for specific datasets. 
     The position on the Critical distance on Figures~\ref{img:mimic-cd-10} and \ref{img:mimic-cd-30} scale denotes test statistic in the Friedman test. Methods that are connected with horizontal lines are statistically indistinguishable.
     }
     \label{img:mimic_logistic}
\end{figure}

To determine the quality of the proposed warm-start, we present on Figure~\ref{img:mimic-adtm} trajectories of ADTM for each cross validation fold with averages and confidence intervals for each considered method. In Figures~\ref{img:mimic-cd-10} and~\ref{img:mimic-cd-30}  we show the statistical significance of our results aggregated across all folds on critical distance plots. 
There is a significant difference in the results of the
specific methods. It can be noticed that the rank method is significantly better than other methods.
Moreover, both encoders are indistinguishable in terms of performance on the 10th and 30th iteration. Corresponding plots for UCI are presented in Appendix \ref{app:uci}. In all cases, baselines either were statistically undistinguishable from encoder-based approaches or even outperformed them. \textbf{This contradicts the assumption that the similarity of general datasets' representations implies that they share the best HP configurations.}


\section{Conclusion}
In our study, we provide results showing that using encoder-based representations has no significant gain over simpler methods in the task of choosing warm-start points in BO. Moreover, our results show that meta-learning methods do not necessarily outperform simple random initialization. To make the evidence stronger, we propose a new method to extract data representation from the model proposed in \citep{iwata_meta-learning_2020}. We evaluate its performance and compare it to Dataset2Vec, an established dataset encoder. Both of them met Requirement~\ref{ass:similarity}. The experiments show that they have comparable results on most warm-start tasks and both of them yield no significant improvement in considered meta-task.

\section{Limitation and Broader Impact}

Here, we want to emphasize some of the limitations of our work. Additionally, we summarize the potential future directions in this field.

We only used two encoders because few can work on heterogeneous data. However, considering the benchmark provided in \citep{zhu_tabular_2023}, we choose the two best models that work on varying features and target spaces. To perform experiments, we used three significantly different sets of datasets. 
Despite their versatility, there is a need to re-perform experiments on a more extensive data portfolio to ensure independence of the results and the data. Concerning reproducibility, one of the sets cannot be published (metaMIMIC) but detailed instructions to reproduce are provided in original paper~\citep{woznica_consolidated_2023}.

We believe that creating new heuristics and encoders is the next direction in this study. The results shown in this paper suggest that this field needs to be explored more. In particular, there is a great need to create encoders or heuristic-based methods that perform in all tasks better than random sampling starting points. \\
\newline
\noindent
\textbf{Social impact statement}

\noindent
After careful reflection, we have determined that this work presents no notable negative impacts on society or the environment.

\bibliography{references_auto}

\section*{Submission Checklist}

\begin{enumerate}
\item For all authors\dots
  \begin{enumerate}
  \item Do the main claims made in the abstract and introduction accurately
    reflect the paper's contributions and scope?
    \answerYes{}
  \item Did you describe the limitations of your work?
    \answerYes{}
  \item Did you discuss any potential negative societal impacts of your work?
    \answerYes{}
  \item Did you read the ethics review guidelines and ensure that your paper
    conforms to them? \url{https://2022.automl.cc/ethics-accessibility/}
    \answerYes{}
  \end{enumerate}
\item If you ran experiments\dots
  \begin{enumerate}
  \item Did you use the same evaluation protocol for all methods being compared (e.g.,
    same benchmarks, data (sub)sets, available resources)?
    \answerYes{}
  \item Did you specify all the necessary details of your evaluation (e.g., data splits,
    pre-processing, search spaces, hyperparameter tuning)?
    \answerYes{}
  \item Did you repeat your experiments (e.g., across multiple random seeds or splits) to account for the impact of randomness in your methods or data?
    \answerYes{}
  \item Did you report the uncertainty of your results (e.g., the variance across random seeds or splits)?
    \answerYes{}
  \item Did you report the statistical significance of your results?
    \answerYes{}
  \item Did you use tabular or surrogate benchmarks for in-depth evaluations?
    \answerYes{}
  \item Did you compare performance over time and describe how you selected the maximum duration?
    \answerNo{The methods presented in this paper are meant to be significantly faster than single evaluation in Bayes Optimization and thus, we do not set time but iteration limit.}
  \item Did you include the total amount of compute and the type of resources
    used (e.g., type of \textsc{gpu}s, internal cluster, or cloud provider)?
    \answerYes{}
  \item Did you run ablation studies to assess the impact of different
    components of your approach?
    \answerNo{We do not propose any new method in this paper except simple heuristic that are too simple to perform ablation studies on them.}
  \end{enumerate}
\item With respect to the code used to obtain your results\dots
  \begin{enumerate}
\item Did you include the code, data, and instructions needed to reproduce the
    main experimental results, including all requirements (e.g.,
    \texttt{requirements.txt} with explicit versions), random seeds, an instructive
    \texttt{README} with installation, and execution commands (either in the
    supplemental material or as a \textsc{url})?
    \answerNo{We fulfilled all requirements from this points except providing all data, as metaMIMIC dataset cannot be published.}
  \item Did you include a minimal example to replicate results on a small subset
    of the experiments or on toy data?
    \answerYes{}
  \item Did you ensure sufficient code quality and documentation so that someone else
    can execute and understand your code?
    \answerYes{}
  \item Did you include the raw results of running your experiments with the given
    code, data, and instructions?
    \answerYes{}
  \item Did you include the code, additional data, and instructions needed to generate
    the figures and tables in your paper based on the raw results?
    \answerYes{}
  \end{enumerate}
\item If you used existing assets (e.g., code, data, models)\dots
  \begin{enumerate}
  \item Did you cite the creators of used assets?
    \answerYes{}
  \item Did you discuss whether and how consent was obtained from people whose
    data you're using/curating if the license requires it?
    \answerYes{}
  \item Did you discuss whether the data you are using/curating contains
    personally identifiable information or offensive content?
    \answerYes{}
  \end{enumerate}
\item If you created/released new assets (e.g., code, data, models)\dots
  \begin{enumerate}
    \item Did you mention the license of the new assets (e.g., as part of your code submission)?
    \answerNA{}
    \item Did you include the new assets either in the supplemental material or as
    a \textsc{url} (to, e.g., GitHub or Hugging Face)?
    \answerNA{}
  \end{enumerate}
\item If you used crowdsourcing or conducted research with human subjects\dots
  \begin{enumerate}
  \item Did you include the full text of instructions given to participants and
    screenshots, if applicable?
    \answerNA{}
  \item Did you describe any potential participant risks, with links to
    Institutional Review Board (\textsc{irb}) approvals, if applicable?
    \answerNA{}
  \item Did you include the estimated hourly wage paid to participants and the
    total amount spent on participant compensation?
    \answerNA{}
  \end{enumerate}
\item If you included theoretical results\dots
  \begin{enumerate}
  \item Did you state the full set of assumptions of all theoretical results?
    \answerNA{}
  \item Did you include complete proofs of all theoretical results?
    \answerNA{}
  \end{enumerate}
\end{enumerate}

\newpage
\appendix
\section{Liltab's representations of datasets}
\label{app:liltab}

Here, we describe representations generated by the \emph{liltab} which is part of the architecture introduced \citep{iwata_meta-learning_2020}, namely the inference network. Please note that in the original paper, there are three parts of the proposed network while we show only the first two of them. This is because the second part creates a representation for each of the records in the dataset. On the contrary, the representations produced by the third part are created attribute-wise and as a result, are harder to interpret. In the original paper, the whole architecture was designed to serve in few-shot learning tasks. However, we do not use any of the few-shot learning features in our work so we do not use terms related to this technique in the description below.  

\subsection{Network architecture}
Let's denote $[X,Y]$ as task, $x_n$ as $n$-th observation in it and $x_{ni}$ as $i$-th attribute of $n$-th observation. In our research, we focus only on the binary target so $y_n\in\{0,1\}$ denotes $n$-th target value. We use here $N$ as the number of observations in the task and $I$ as the number of attributes so $n=1,\dots,N$ and $i=1,\dots,I$. 

Let's denote feed-forward neural networks as $f_\odot$ when their function is to encode information about attributes and targets and $g_\odot$ when their function is to aggregate the outputs of $f_\odot$ networks. Vector $\bar{\nu}_i$ and scalar $\bar{c}_i$ store information about the marginal distribution of each attribute and a target. Their values are obtained by feed forward networks $f_{\overline{\nu}}$, $f_{\overline{c}}$, $g_{\overline{\nu}}$, $g_{\overline{c}}$, which is presented in Figure \ref{img:inference-1}. The formal formulation of this step is presented below:

\begin{equation}
\label{eq:step-1}
    \bar{\nu}_i = g_{\bar{\nu}}\left(\frac{1}{N}\sum\limits^N_{n=1}f_{\bar{\nu}}(x_{ni})\right)
    \text{,\quad}
    \bar{c} = g_{\bar{c}}\left(\frac{1}{N}\sum\limits^N_{n=1}f_{\bar{c}}(y_n)\right)
    \text{.}
\end{equation}

\vspace{0.5cm}
\begin{figure}[h]
    \centering
    \includegraphics[width=\textwidth]{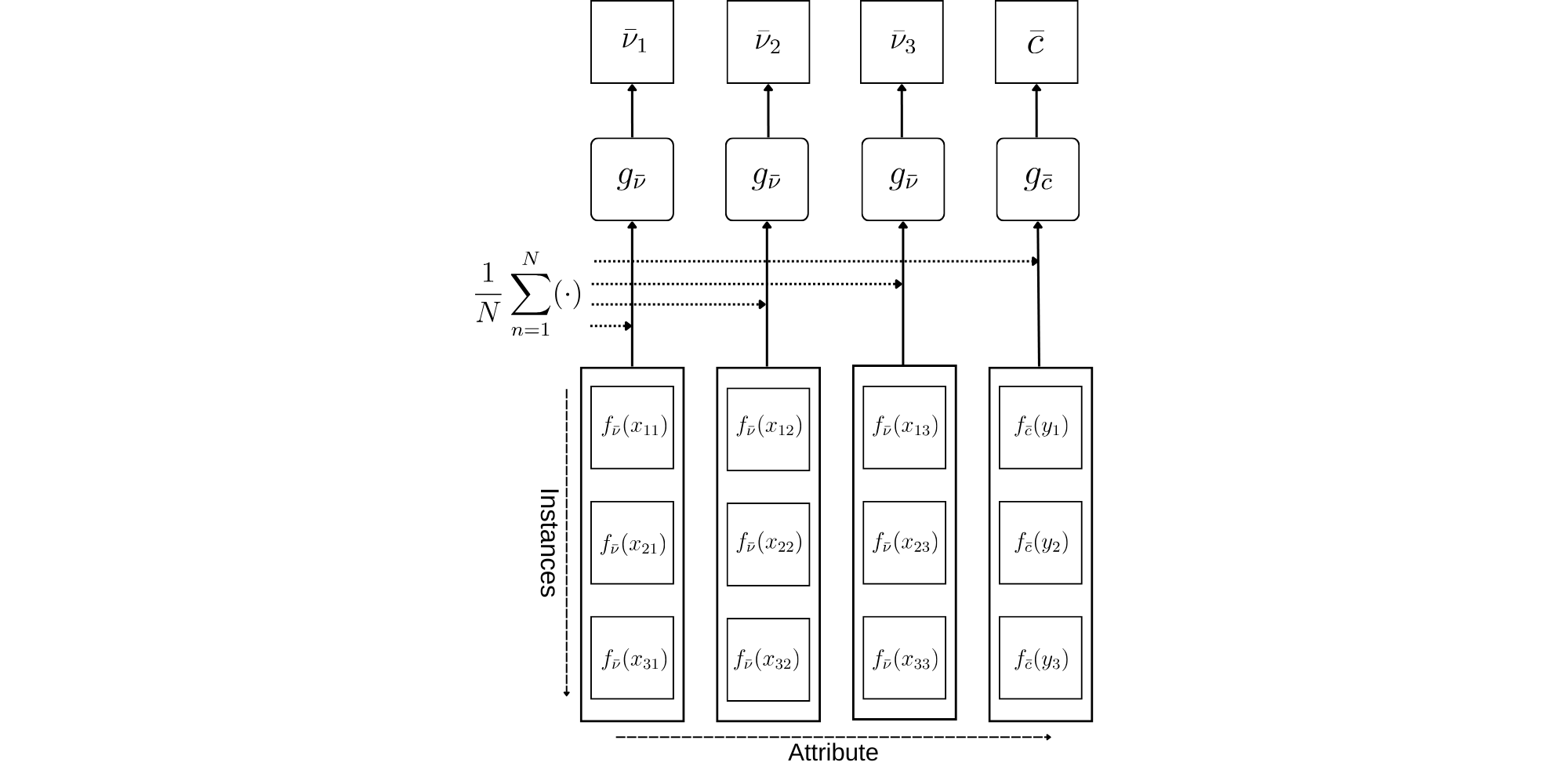}
    \caption{First step of the inference network. Here, the inference network learns about the~empirical marginal distributions of the attributes based on the support set.}
    \label{img:inference-1}
\end{figure}

\begin{figure}[h]
    \centering
    \includegraphics[width=\textwidth]{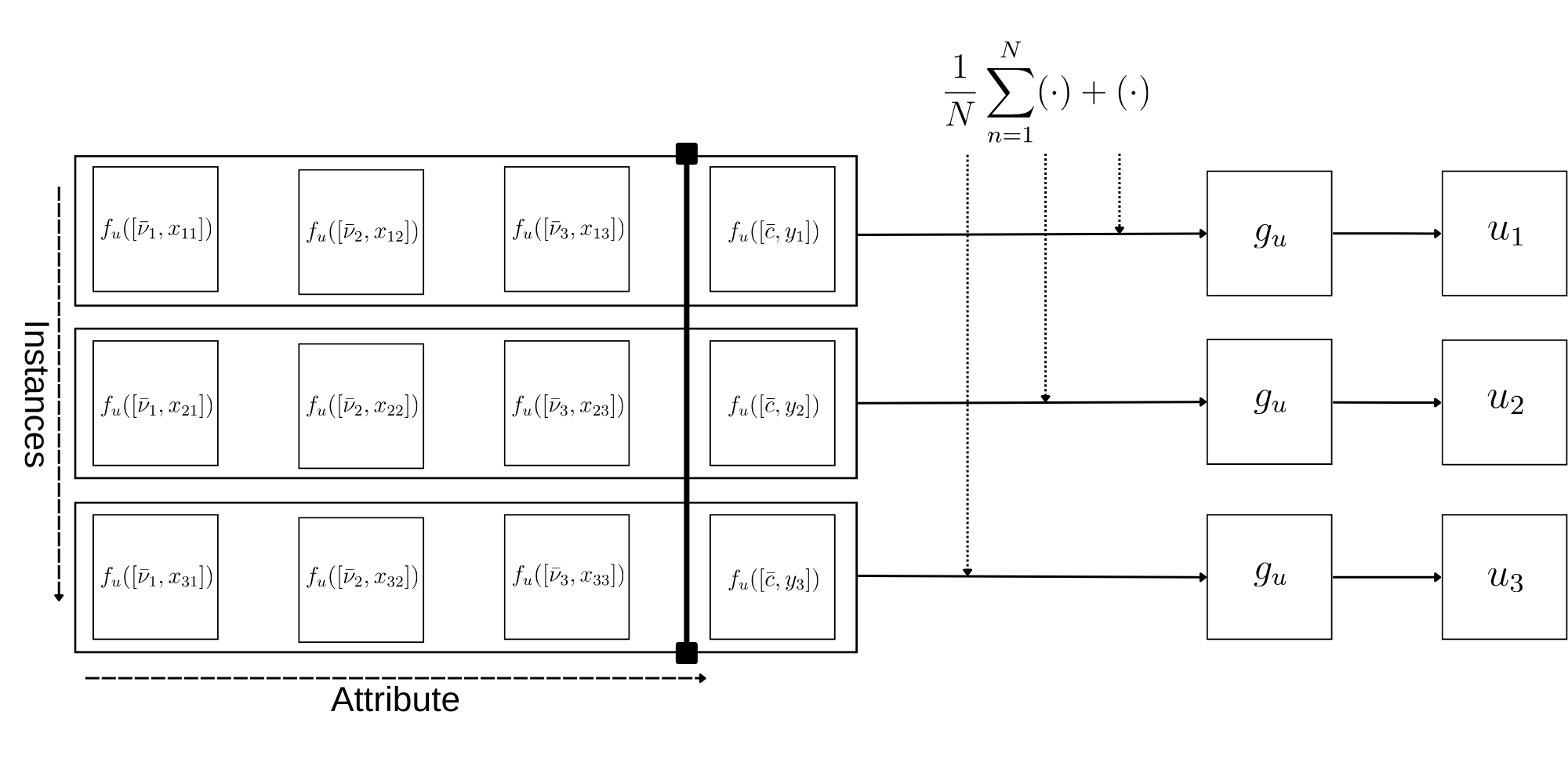}
    \caption{Second step of the inference network. Here, the inference network learns the relationships between the attributes and the target based on the support set.}
    \label{img:inference-2}
\end{figure}

Further encoding $u_n$ provides the possibility to provide additional information about relationships between attributes and targets. What is more, this representation is created for each observation (as shown in Figure \ref{img:inference-2}). This step can be formulated as below:

\begin{equation}
\label{eq:step-2}
    u_n = g_u\left(\frac{1}{I}\sum\limits^I_{i=1}f_u([\bar{\nu_i},x_{ni}])+f_u([\bar{c},y_n])\right)
    \text{,}
\end{equation}

where $[\cdot]$ denotes the concatenation. Please note that the representations $u=(u_1,\dots,u_N)$ is in $\mathbb{R}^{N\times p}$, where $p$ denotes the size of representation and is treated as the architecture's hyperparameter. The presented architecture can store information about marginal distributions and interactions between variables in the dataset that is provided on input. Thus, by providing an appropriate loss function, one can train such a model to distinguish between different datasets or subsets of datasets.

\vspace{0.5cm}

\subsection{Network training}
\noindent During the original training procedure, the authors of \citep{iwata_meta-learning_2020} wanted to maximize its predictive performance. However, representations produced by an inference network occur to be uninformative on real-world data and often are vectors with few non-zero coordinates. To overcome that issue, we propose a new encoder-based technique that targets this part of the model. 

In our training procedure, each batch of data is created from samples from several different datasets, where the numbers of samples for each dataset are equal. Next, observations from one dataset in the batch are encoded to their representation.
During the calculation of the loss function, we focus on representations of specific observations and train the network to produce encodings that are close to each other when they correspond to the same dataset and far away when they relate to different datasets. To achieve that, we propose the contrastive learning approach with loss function inspired by \citep{song_deep_2015}:

\begin{equation*}
    loss = \frac{1}{|\mathcal{N}| + |\mathcal{S}|}\left(\sum_{i, j \in \mathcal{S}} \|\phi(x_i) - \phi(x_j)\|_2 - \sum_{i, j \in \mathcal{N}} \|\phi(x_i) - \phi(x_j)\|_2\right) \text{,}
\end{equation*}

\noindent
where $x_i$ denotes specific observation, $\phi$ is a \textit{liltab}'s encoder function, $\mathcal{S}$ is a set of pairs of indices of observations that belong to the same dataset, and $\mathcal{N}$ denotes pairs of indices of observations that are from different datasets. Note that we index by observations, not by subsets in batch.

\section{Data}
\label{app:data}

The set of datasets from OpenML are binary classification tasks created from regression ones when a classification target is obtained as an indicator of whether the regression target is above or below the average. The datasets are preprocessed and standardized, so the distribution of ones and zeros is equal. A complete list of these datasets is in Appendix \ref{app:openml_data}.

The next set of data is supplied from the GitHub repository of D2V. Some of the datasets are multilabel classification tasks with several classes varying from 2 to 10. However, we decided to cast classes so that there are only 2 classes in every task. The reason for such modification is that the tasks on which the encoders are trained are binary only. We also remove datasets with more than 10000 observations or 20 features. Appendix \ref{app:uci_data} shows a complete list of used datasets.

The last set of datasets is prepared on the metaMIMIC datasets \citep{woznica_consolidated_2023}. It contains fully preprocessed data from \citep{johnson_mimic-iv_2023}, with 12 binary classification tasks, each describing the presence of a specific disease. As tasks originally contain more than 100 features, selecting the 10 most important for each task provides the ability to measure the similarity of tasks by simply calculating the fraction of shared most significant features. During our study, we would like to consider several small datasets, as it is the preferred data type for \textit{liltab}. Due to that fact, we split each metaMIMIC task into multiple smaller tasks.  For every small task used in the warm-start evaluation, we specify 4 observations as the training set and 29 observations as the validation set. Both training and validation sets are sampled equally concerning the target value. With this methodology, we yielded over 5000 small tasks in total. MetaMIMIC has not yet been used in this domain, especially as the source of small subsets of data. However, it is a promising type of set of tasks for the dataset similarity as each task has some common features with the others.

\newpage

\section{Datasets from OpenML}
\label{app:openml_data}
\begin{longtable}{lrr}
\caption{Datasets from OpenML used for encoder training.}
\label{tab:openml_data} \\
\toprule
name & number of rows & number of columns \\
\midrule
\endfirsthead
\toprule
name & number of rows & number of columns \\
\midrule
\endhead
\endfoot
\bottomrule
\endlastfoot
\texttt{diggle\_table\_a1} & 48 & 5 \\ \hline
\texttt{fri\_c4\_100\_25} & 100 & 26 \\ \hline
\texttt{chatfield\_4} & 235 & 13 \\ \hline
\texttt{edm} & 154 & 18 \\ \hline
\texttt{vineyard} & 52 & 3 \\ \hline
\texttt{pollution} & 60 & 16 \\ \hline
\texttt{fri\_c3\_100\_10} & 100 & 11 \\ \hline
\texttt{visualizing\_hamster} & 73 & 6 \\ \hline
\texttt{pyrim} & 74 & 28 \\ \hline
\texttt{visualizing\_slope} & 44 & 4 \\ \hline
\texttt{hip} & 54 & 8 \\ \hline
\texttt{chscase\_geyser1} & 222 & 3 \\ \hline
\texttt{rabe\_148} & 66 & 6 \\ \hline
\texttt{bodyfat} & 252 & 15 \\ \hline
\texttt{chscase\_vine1} & 52 & 10 \\ \hline
\texttt{sleuth\_ex1714} & 47 & 8 \\ \hline
\texttt{echocardiogram-uci} & 132 & 8 \\ \hline
\texttt{hutsof99\_child\_witness} & 42 & 16 \\ \hline
\texttt{fri\_c3\_250\_10} & 250 & 11 \\ \hline
\texttt{chscase\_funds} & 185 & 2 \\ \hline
\texttt{qsartox} & 16 & 24 \\ \hline
\texttt{qsabr2} & 15 & 10 \\ \hline
\texttt{sleep} & 62 & 8 \\ \hline
\texttt{longley} & 16 & 7 \\ \hline
\texttt{heart} & 270 & 14 \\ \hline
\texttt{mu284} & 284 & 10 \\ \hline
\texttt{ICU} & 200 & 20 \\ \hline
\texttt{qsbralks} & 13 & 22 \\ \hline
\texttt{analcatdata\_uktrainacc} & 31 & 16 \\ \hline
\texttt{slump} & 103 & 10 \\ \hline
\texttt{rabe\_131} & 50 & 6 \\ \hline
\texttt{gascons} & 27 & 5 \\ \hline
\texttt{rabe\_265} & 51 & 7 \\ \hline
\texttt{rabe\_166} & 40 & 2 \\ \hline
\texttt{baskball} & 96 & 5 \\ \hline
\texttt{bolts} & 40 & 7 \\ \hline
\texttt{chscase\_demand} & 27 & 11 \\ \hline
\texttt{qsfsr2} & 19 & 10 \\ \hline
\texttt{fri\_c2\_100\_5} & 100 & 6 \\ \hline
\texttt{sleuth\_ex1605} & 62 & 6 \\ \hline
\texttt{qsfsr1} & 20 & 10 \\ \hline
\texttt{visualizing\_ethanol} & 88 & 3 \\
\texttt{EgyptianSkulls} & 150 & 5 \\ \hline
\texttt{transplant} & 131 & 3 \\ \hline
\texttt{pwLinear} & 200 & 11 \\ \hline
\texttt{diabetes\_numeric} & 43 & 3 \\ \hline
\texttt{autoPrice} & 159 & 16 \\ \hline
\texttt{treepipit} & 86 & 10 \\ \hline
\texttt{branin} & 225 & 3 \\ \hline
\texttt{machine\_cpu} & 209 & 7 \\ \hline
\texttt{detroit} & 13 & 14 \\ \hline
\texttt{fri\_c2\_250\_25} & 250 & 26 \\ \hline
\texttt{qsbr\_y2} & 25 & 10 \\ \hline
\texttt{humans\_numeric} & 75 & 15 \\ \hline
\texttt{visualizing\_environmental} & 111 & 4 \\ \hline
\texttt{USCrime} & 47 & 14 \\ \hline
\texttt{rabe\_176} & 70 & 4 \\ \hline
\texttt{fri\_c0\_250\_5} & 250 & 6 \\ \hline
\texttt{rabe\_266} & 120 & 3 \\
\end{longtable}

\newpage

\section{Datasets from UCI}
\label{app:uci_data}
\begin{longtable}{lrr}
\caption{Datasets from UCI used for HPO warmstart evaluation.}
\label{tab:uci_data} \\
\toprule
name & number of rows & number of columns \\
\midrule
\endfirsthead
\toprule
name & number of rows & number of columns \\
\midrule
\endhead
\endfoot
\bottomrule
\endlastfoot
\texttt{chess-krvkp} & 3196 & 37 \\ \hline
\texttt{post-operative} & 90 & 9 \\ \hline
\texttt{contrac} & 1473 & 10 \\ \hline
\texttt{pittsburg-bridges-REL-L} & 103 & 8 \\ \hline
\texttt{seeds} & 210 & 8 \\ \hline
\texttt{ecoli} & 336 & 8 \\ \hline
\texttt{acute-nephritis} & 120 & 7 \\ \hline
\texttt{haberman-survival} & 306 & 4 \\ \hline
\texttt{cylinder-bands} & 512 & 36 \\ \hline
\texttt{pittsburg-bridges-TYPE} & 105 & 8 \\ \hline
\texttt{ringnorm} & 7400 & 21 \\ \hline
\texttt{tic-tac-toe} & 958 & 10 \\ \hline
\texttt{led-display} & 1000 & 8 \\ \hline
\texttt{hill-valley} & 1212 & 101 \\ \hline
\texttt{waveform} & 5000 & 22 \\ \hline
\texttt{credit-approval} & 690 & 16 \\ \hline
\texttt{dermatology} & 366 & 35 \\ \hline
\texttt{statlog-heart} & 270 & 14 \\ \hline
\texttt{echocardiogram} & 131 & 11 \\ \hline
\texttt{thyroid} & 7200 & 22 \\ \hline
\texttt{planning} & 182 & 13 \\ \hline
\texttt{spect} & 265 & 23 \\ \hline
\texttt{musk-1} & 476 & 167 \\ \hline
\texttt{wall-following} & 5456 & 25 \\ \hline
\texttt{ozone} & 2536 & 73 \\ \hline
\texttt{hayes-roth} & 160 & 4 \\ \hline
\texttt{breast-cancer-wisc-prog} & 198 & 34 \\ \hline
\texttt{yeast} & 1484 & 9 \\ \hline
\texttt{energy-y2} & 768 & 9 \\ \hline
\texttt{wine-quality-red} & 1599 & 12 \\ \hline
\texttt{teaching} & 151 & 6 \\ \hline
\texttt{musk-2} & 6598 & 167 \\ \hline
\texttt{zoo} & 101 & 17 \\ \hline
\texttt{statlog-landsat} & 6435 & 37 \\ \hline
\texttt{molec-biol-promoter} & 106 & 58 \\ \hline
\texttt{heart-switzerland} & 123 & 13 \\ \hline
\texttt{congressional-voting} & 435 & 17 \\ \hline
\texttt{statlog-image} & 2310 & 19 \\ \hline
\texttt{oocytes\_trisopterus\_states\_5b} & 912 & 33 \\ \hline
\texttt{oocytes\_merluccius\_nucleus\_4d} & 1022 & 42 \\ \hline
\texttt{wine-quality-white} & 4898 & 12 \\ \hline
\texttt{mushroom} & 8124 & 22 \\
\texttt{conn-bench-sonar-mines-rocks} & 208 & 61 \\ \hline
\texttt{vertebral-column-3clases} & 310 & 7 \\ \hline
\texttt{heart-hungarian} & 294 & 13 \\ \hline
\texttt{low-res-spect} & 531 & 101 \\ \hline
\texttt{semeion} & 1593 & 257 \\ \hline
\texttt{iris} & 150 & 5 \\ \hline
\texttt{ilpd-indian-liver} & 583 & 10 \\ \hline
\texttt{optical} & 5620 & 63 \\ \hline
\texttt{horse-colic} & 368 & 26 \\ \hline
\texttt{waveform-noise} & 5000 & 41 \\ \hline
\texttt{lenses} & 24 & 5 \\ \hline
\texttt{glass} & 214 & 10 \\ \hline
\texttt{mammographic} & 961 & 6 \\ \hline
\texttt{twonorm} & 7400 & 21 \\ \hline
\texttt{balance-scale} & 625 & 5 \\ \hline
\texttt{abalone} & 4177 & 9 \\ \hline
\texttt{pittsburg-bridges-T-OR-D} & 102 & 8 \\ \hline
\texttt{bank} & 4521 & 17 \\ \hline
\texttt{hepatitis} & 155 & 20 \\ \hline
\texttt{breast-cancer-wisc} & 699 & 10 \\ \hline
\texttt{ionosphere} & 351 & 34 \\ \hline
\texttt{flags} & 194 & 29 \\ \hline
\texttt{image-segmentation} & 2310 & 19 \\ \hline
\texttt{breast-tissue} & 106 & 10 \\ \hline
\texttt{monks-2} & 601 & 7 \\ \hline
\texttt{pittsburg-bridges-SPAN} & 92 & 8 \\ \hline
\texttt{trains} & 10 & 30 \\ \hline
\texttt{heart-cleveland} & 303 & 14 \\ \hline
\texttt{spectf} & 267 & 45 \\ \hline
\texttt{page-blocks} & 5473 & 11 \\ \hline
\texttt{statlog-vehicle} & 846 & 19 \\ \hline
\texttt{monks-3} & 554 & 7 \\ \hline
\texttt{blood} & 748 & 5 \\ \hline
\texttt{oocytes\_merluccius\_states\_2f} & 1022 & 26 \\ \hline
\texttt{statlog-australian-credit} & 690 & 15 \\ \hline
\texttt{energy-y1} & 768 & 9 \\ \hline
\texttt{heart-va} & 200 & 13 \\ \hline
\texttt{steel-plates} & 1941 & 28 \\ \hline
\texttt{breast-cancer} & 286 & 10 \\ \hline
\texttt{lung-cancer} & 32 & 57 \\ \hline
\texttt{wine} & 178 & 14 \\ \hline
\texttt{spambase} & 4601 & 58 \\ \hline
\texttt{oocytes\_trisopterus\_nucleus\_2f} & 912 & 26 \\ \hline
\texttt{pima} & 768 & 9 \\ \hline
\texttt{annealing} & 898 & 32 \\ \hline
\texttt{breast-cancer-wisc-diag} & 569 & 31 \\
\texttt{lymphography} & 148 & 19 \\ \hline
\texttt{car} & 1728 & 7 \\ \hline
\texttt{cardiotocography-10clases} & 2126 & 22 \\ \hline
\texttt{titanic} & 2201 & 4 \\ \hline
\texttt{acute-inflammation} & 120 & 7 \\ \hline
\texttt{pittsburg-bridges-MATERIAL} & 106 & 8 \\ \hline
\texttt{monks-1} & 556 & 7 \\ \hline
\texttt{fertility} & 100 & 10 \\ \hline
\texttt{balloons} & 16 & 5 \\ \hline
\texttt{vertebral-column-2clases} & 310 & 7 \\ \hline
\texttt{synthetic-control} & 600 & 61 \\ \hline
\texttt{parkinsons} & 195 & 23 \\ \hline
\texttt{statlog-german-credit} & 1000 & 25 \\ \hline
\texttt{molec-biol-splice} & 3190 & 61 \\ \hline
\texttt{cardiotocography-3clases} & 2126 & 22 \\
\end{longtable}

\newpage

\section{Hyperparameters of encoders}
\label{app:encoders_hp}
\begin{table}[h]
\label{tab:liltab_hp}
\caption{Hyperparameters of \textit{liltab} encoder.}
\centering
\begin{tabular}{lll}
\hline
Hyperparameter & Description & Value \\ \hline
num\_epochs & Maximum number of epochs & 100000 \\ \hline
learning\_rate & Learning rate & 0.0001 \\ \hline
weight\_decay & Weight decay & 0 \\ \hline
batch\_size & Batch size & 37 \\ \hline
early\_stopping\_epochs & Epochs without loss decrease before stop of training & 2500 \\ \hline
hidden\_representation\_size & Dimensionality of output of intermediate networks & 32 \\ \hline
n\_hidden\_layers & Number of hidden layers in intermediate networks & 3 \\ \hline
hidden\_size & Size of hidden layers in intermediate networks & 32 \\ \hline
dropout\_rate & Dropout rate in all intermediate networks & 0.1 \\ \hline
\end{tabular}
\end{table}

\begin{table}[h]
\centering
\label{tab:d2v_hp}
\caption{Hyperparameters of Dataset2Vec encoder.}
\begin{tabular}{lll}
\hline
Hyperparameter & Description & Value \\ \hline
gamma & Scaling factor in metric-based classification & 1 \\ \hline
num\_epochs & Maximum number of epochs & 100000 \\ \hline
learning\_rate & Learning rate & 0.001 \\ \hline
weight\_decay & Weight decay & 0.0001 \\ \hline
batch\_size & Batch size & 16 \\ \hline
train\_n\_batches & Number of batches in one epoch & 100 \\ \hline
early\_stopping\_epochs & Epochs without loss decrease before stop of training & 500 \\ \hline
f\_dense\_hidden\_size & Size of input/output layers in residual block in first layer & 16 \\ \hline
f\_res\_hidden\_size & Size of hidden layers in residual block in first layer & 16 \\ \hline
f\_res\_n\_hidden & Number of hidden layers in residual block in first layer & 3 \\ \hline
f\_dense\_out\_hidden\_size & Output size of first layer & 16 \\ \hline
f\_block\_repetitions & Number of residual blocks in first layer & 3 \\ \hline
g\_layers\_sizes & Sizes of hidden layers in second layer & {[}32, 16, 8{]} \\ \hline
h\_dense\_hidden\_size & Size of input/output layers in residual block in third layer & 32 \\ \hline
h\_res\_hidden\_size & Size of hidden layers in residual block in third layer & 32 \\ \hline
h\_res\_n\_hidden & Number of hidden layers in residual block in third layer & 3 \\ \hline
h\_dense\_out\_hidden\_size & Output size of last layer & 16 \\ \hline
h\_block\_repetitions & Number of residual blocks in third layer & 3 \\ \hline
\end{tabular}
\end{table}

\newpage

\section{Hyperparameters grids used for HPO}
\label{app:ho}
\begin{table}[h]
\caption{Hyperparameters grid used for HPO of elastic net algorithm.}
\label{tab:elasticnet_ho}
\centering
\begin{tabular}{llll}
\hline
Condition & Hyperparameter & Values' range & Distribution \\ \hline
n/a & tol & {[}0.0001, 0.001{]} & loguniform \\ \hline
n/a & C & {[}0.0001, 10000{]} & loguniform \\ \hline
n/a & solver & \begin{tabular}[c]{@{}l@{}}{[}lbfgs, liblinear, newton-cg, \\ newton-cholesky, sag, saga{]}\end{tabular} & categorical \\ \hline
\multirow{2}{*}{solver = liblinear} & intercept scaling & {[}0.001, 1{]} & uniform \\ \cline{2-4} 
 & penalty & {[}l1, l2{]} & categorical \\ \hline
\begin{tabular}[c]{@{}l@{}}solver = liblinear\\ and penalty = l2\end{tabular} & dual & {[}true, false{]} & categorical \\ \hline
\multirow{2}{*}{solver = saga} & penalty & {[}elasticnet, l1, l2, null{]} & categorical \\ \cline{2-4} 
 & l1 ratio & {[}0, 1{]} & uniform \\ \hline
\begin{tabular}[c]{@{}l@{}}solver $\neq$ saga\\ and solver $\neq$ liblinear \end{tabular} & penalty & {[}l2, null{]} & categorical \\ \hline
\end{tabular}

\end{table}

\begin{table}[h]
\caption{Hyperparameters grid used for HPO of XGBoost algorithm.}
\label{tab:xgboost_ho}
\centering
\begin{tabular}{lll}
\hline
Hyperparameter & Values' range & Distribution \\ \hline
no of estimators & {[}1, 1000{]} & uniform (int) \\ \hline
learning rate & {[}0, 1{]} & uniform \\ \hline
booster & {[}gblinear, gbtree{]} & categorical \\ \hline
subsample & {[}0.5, 1{]} & uniform \\ \hline
max depth & {[}6, 15{]} & uniform (int) \\ \hline
min child weight & {[}2, 256{]} & uniform \\ \hline
colsample bytree & {[}0.2, 1{]} & uniform \\ \hline
colsample bylevel & {[}0.2, 1{]} & uniform \\ \hline
\end{tabular}
\end{table}

\newpage

\section{Results on the UCI datasets}
\label{app:uci}

\begin{figure}[!h]
     \centering
     \begin{subfigure}[b]{0.45\textwidth}
         \centering
         \includegraphics[width=\textwidth]{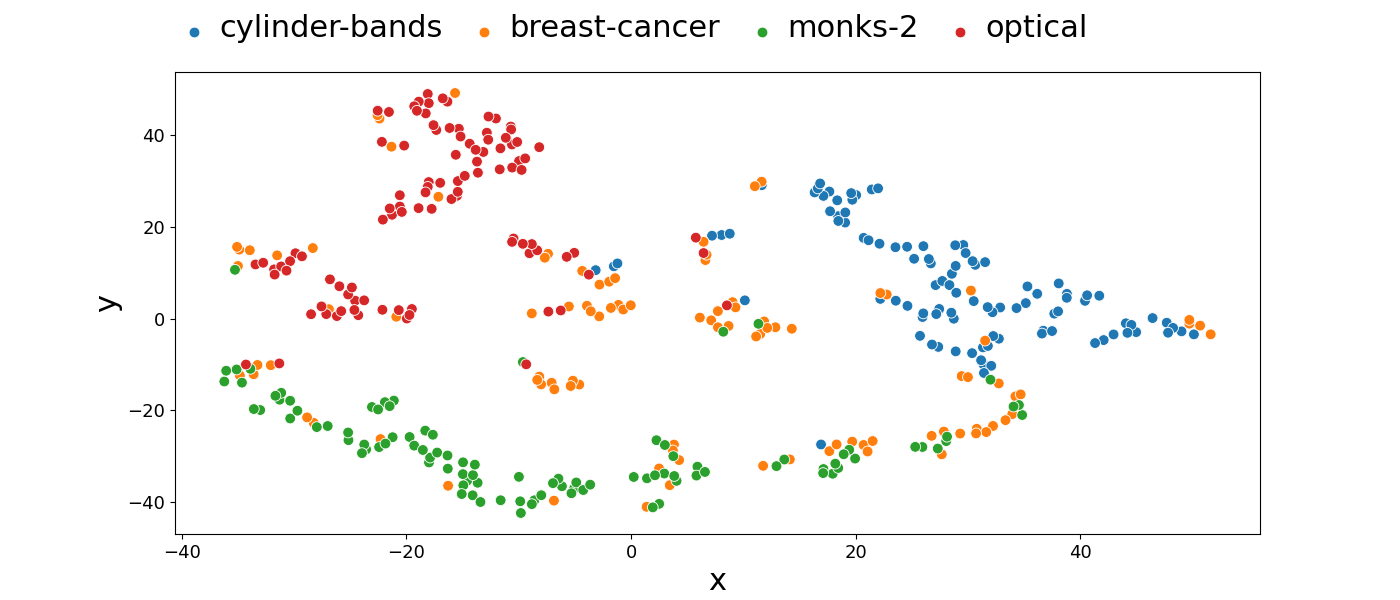}
         \caption{The representation obtained with \textit{liltab}.}
         \label{img:uci-liltab}
     \end{subfigure}
     \hfill
     \begin{subfigure}[b]{0.45\textwidth}
         \centering
         \includegraphics[width=\textwidth]{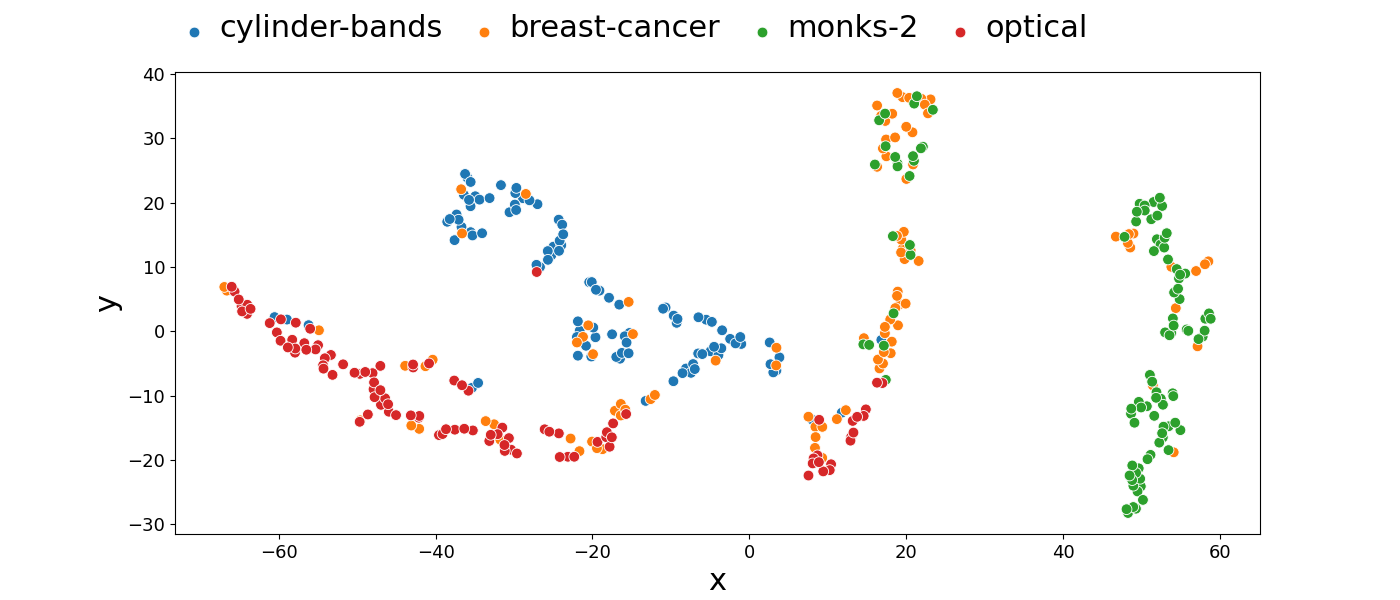}
         \caption{The representation obtained with D2V.}
         \label{img:uci-d2v}
     \end{subfigure}
     \caption{T-SNE visualization of four encoded datasets' representations. Each point represents an encoded subset of data sampled from one of the datasets.}
     \label{img:uci}
\end{figure}

\begin{figure}[!h]
\begin{subfigure}[b]{1\textwidth}
     \centering
     \includegraphics[width=\textwidth]{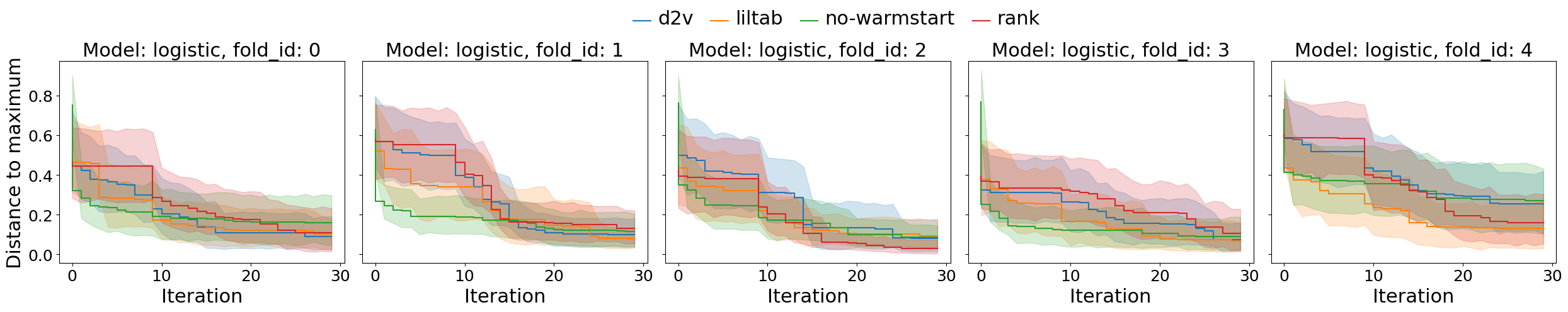}
     \caption{Plots of ADTM of UCI data HPO with warm-start comparison. A lower score means the performance of a method is closer to the best score obtained on a specific dataset on average.}
     \label{img:adtm-uci-logistic}
\end{subfigure}
\hfill
     \centering
     \begin{subfigure}[b]{0.475\textwidth}
         \centering
         \includegraphics[width=\textwidth]{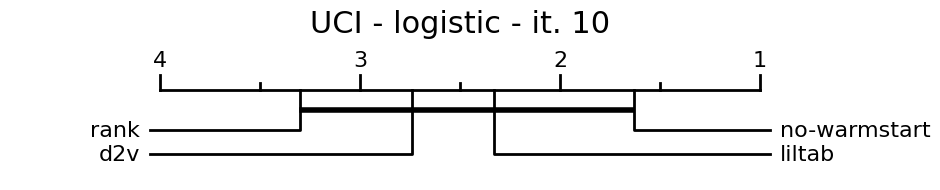}
         \caption{Critical distance plot for logistic regression in 10th iteration.}
         \label{img:cd_uci_logistic_10}
     \end{subfigure}
     \hfill
     \begin{subfigure}[b]{0.475\textwidth}
         \centering
         \includegraphics[width=\textwidth]{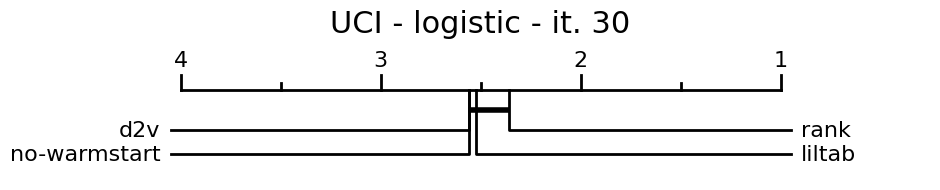}
         \caption{Critical distance plot for logistic regression in 30th iteration.}
        \label{img:cd_uci_logistic_30}
     \end{subfigure}
     \caption{Results for UCI datasets, elastic net. Each fold in ADTM plots on Figure~\ref{img:adtm-uci-logistic} presents results for a specific split of datasets to train and validation the subset. The position on the critical distance scale in Figures \ref{img:cd_uci_logistic_10} and \ref{img:cd_uci_logistic_30} denotes test statistics in the Friedman test. Methods that are connected with horizontal lines are statistically indistinguishable. It shows that the difference in performance between methods is statistically irrelevant.}
     \label{img:uci-logistic}
\end{figure}

\clearpage

\begin{figure}[!h]
\begin{subfigure}[b]{1\textwidth}
     \centering
     \includegraphics[width=\textwidth]{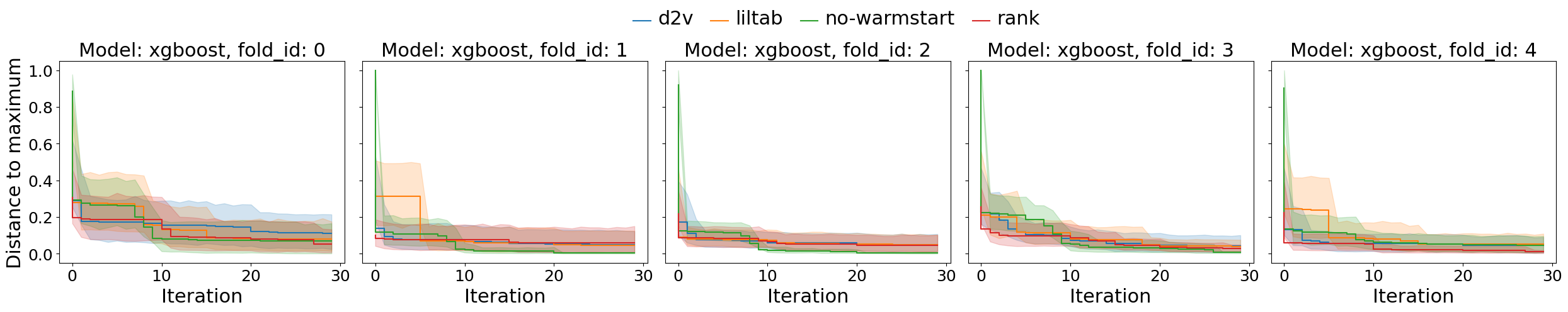}
     \caption{Plots of ADTM of UCI data HPO with warm-start comparison. A lower score means the performance of a method is closer to the best score obtained on a specific dataset on average.}
     \label{img:adtm-uci-xgboost}
\end{subfigure}
\hfill
\centering
     \begin{subfigure}[b]{0.475\textwidth}
         \centering
         \includegraphics[width=\textwidth]{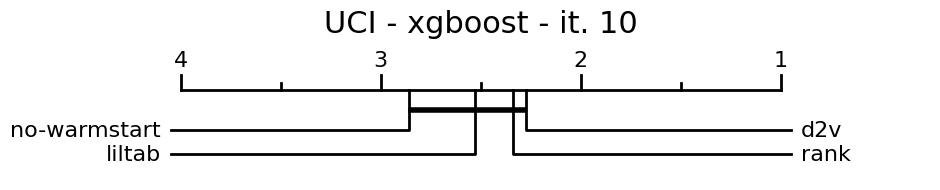}
         \caption{Critical distance plot for XGBoost in 10th iteration.}
         \label{img:uci_xgboost_10}
     \end{subfigure}
     \hfill
     \begin{subfigure}[b]{0.475\textwidth}
         \centering
         \includegraphics[width=\textwidth]{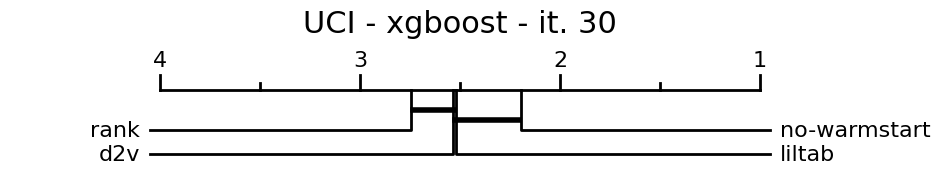}
         \caption{Critical distance plot for XGBoost in 30th iteration.}
         \label{img:uci_xgboost_30}
     \end{subfigure}
     \caption{Results for UCI, XGBoost. Each fold in ADTM plots on Figure~\ref{img:adtm-uci-xgboost} presents results for a specific split of datasets to train and validation the subset. The position on the critical distance scale in Figures \ref{img:uci_xgboost_10} and \ref{img:uci_xgboost_30} denotes test statistics in the Friedman test. Methods that are connected with horizontal lines are statistically indistinguishable. It shows that the difference in performance between methods is statistically irrelevant.}
     \label{img:uci_xgboost}
\end{figure}

\end{document}